\documentclass[11pt,a4paper]{article}
\usepackage[hyperref]{emnlp-ijcnlp-2019}
\usepackage{times}
\usepackage{latexsym}

\usepackage{url}
\usepackage{bm}
\usepackage{graphicx}
\usepackage{multirow, booktabs}
\usepackage{amsmath}
\usepackage{amssymb}
\usepackage{amsfonts}
\usepackage{makecell}
\usepackage{verbatim}
\usepackage{array}
\usepackage{xcolor}
\usepackage{soul}
\usepackage{subfigure}

\aclfinalcopy 

\newcolumntype{L}[1]{>{\raggedright\let\newline\\\arraybackslash\hspace{0pt}}m{#1}}

\usepackage{pifont}
\newcommand{\xmark}{\ding{55}}

\title{Transferable End-to-End Aspect-based Sentiment Analysis with \\ Selective Adversarial Learning}

\author{Zheng Li$^{1}$, Xin Li$^{2}$, Ying Wei$^{3}$, Lidong Bing$^{4}$, Yu Zhang$^{5}$, Qiang Yang$^{1}$ \\
$^{1}$The Hong Kong University of Science and Technology, Hong Kong\\
$^{2}$The Chinese University of Hong Kong, Hong Kong\\
$^{3}$Tencent AI Lab, Shenzhen, China\\
$^{4}$R\&D Center Singapore, Machine Intelligence Technology, Alibaba DAMO Academy \\
$^{5}$Southern University of Science and Technology, Shenzhen, China\\
 {\{zlict, qyang\}@cse.ust.hk, lixin@se.cuhk.edu.hk, judywei@tencent.com} \\
 {l.bing@alibaba-inc.com, yu.zhang.ust@gmail.com}
}

\date{}

\begin{document}
\maketitle
\begin{abstract}
Joint extraction of aspects and sentiments can be effectively formulated as a sequence labeling problem. However, such formulation hinders the effectiveness of supervised methods due to the lack of annotated sequence data in many domains. To address this issue, we firstly explore an unsupervised domain adaptation setting for this task. Prior work can only use common syntactic relations between aspect and opinion words to bridge the domain gaps, which highly relies on external linguistic resources. To resolve it, we propose a novel Selective Adversarial Learning (SAL) method to align the inferred correlation vectors that automatically capture their latent relations. The SAL method can dynamically learn an alignment weight for each word such that more important words can possess higher alignment weights to achieve fine-grained (word-level) adaptation. Empirically, extensive experiments\footnote{The code is available at~\url{https://github.com/hsqmlzno1/Transferable-E2E-ABSA}} demonstrate the effectiveness of the proposed SAL method.
\end{abstract}

\section{Introduction}
End-to-End Aspect-Based Sentiment Analysis (E2E-ABSA) aims to jointly detect the aspect terms explicitly mentioned in sentences and predict the sentiment polarities over them~\cite{liu2012sentiment,pontiki2014semeval}. For example, in the sentence ``{\it The \textbf{AMD Turin Processor} seems to always perform much better than {\it \textbf{Intel}}}'', the user mentions two aspect terms, i.e., ``{\it \textbf{AMD Turin Processor}}'' and ``{\it \textbf{Intel}}'', and expresses positive and negative sentiments over them, respectively. 

Typically, prior work formulates E2E-ABSA as a sequence labeling problem over a unified tagging scheme~\cite{mitchell2013open,zhang2015neural,li2019unified}. The unified tagging scheme connects a set of {\it aspect boundary tags} (e.g., \{\texttt{B}, \texttt{I}, \texttt{E}, \texttt{S}, \texttt{O}\} denotes the beginning of, inside of, end of, single-word, and no aspect term), and {\it sentiment tags} (e.g. \{\texttt{POS}, \texttt{NEG}, \texttt{NEU}\} denotes positive, negative or neutral sentiment) together to constitute a joint label space for each word. As such, ``{\it \textbf{AMD Turin Processor}}'' and ``{\it \textbf{Intel}}'' should be tagged with \{\texttt{B-POS}, \texttt{I-POS}, \texttt{E-POS}\} and \{\texttt{S-NEG}\}, respectively, while the remaining words are tagged with \texttt{O}. This formulation makes two sub-tasks joint modeling easier, and meanwhile, tend to be low-resource. There usually exist few annotated data for each new domain, where labeling each word with a unified tag could be more time-consuming and expensive.

To alleviate the dependence on domain supervisions, we explore an unsupervised domain adaptation setting for E2E-ABSA, which aims to leverage knowledge from a labeled source domain to improve the sequence learning in an unlabeled target domain. The challenges in fulfillment of this setting are two-fold: (1) there exists a large feature distribution shift between domains since aspect terms in different domains are usually disjoint. For example, users usually mention ``{\it \textbf{pizza}}'' in the {\it Restaurant} domain while ``{\it {\textbf{camera}}}'' is often discussed in the {\it Laptop} domain; (2) Unlike domain adaptation in traditional sentiment classification~\cite{blitzer2007biographies} that learns shared sentence or document representations, we need to learn fine-grained (word-level) representations to be domain-invariant for sequence prediction. 

Consider the first problem, i.e., what to transfer? Even though aspect terms from different domains behave distinctly, some association patterns between aspect and opinion words are common across domains; e.g., ``{\it The \textbf{pizza} is great.}'' from the {\it Restaurant} domain and ``{\it The \textbf{camera} is excellent.}'' from the {\it Laptop} domain. Both of them share the same syntactic pattern ({\it {aspect words}}$\rightarrow$$nsubj$$\rightarrow${\it opinion words}). Inspired by this, existing studies use general syntactic relations as the pivot to bridge the domain gaps for cross-domain aspect extraction~\cite{jakob2010extracting,ding2017recurrent}, or aspect and opinion co-extraction~\cite{li2012cross,wang2018recursive}. Unfortunately, these methods highly rely on prior knowledge (e.g., manually-designed rules) or external linguistic resources (e.g., dependency parsers), which are inflexible and prone to bringing in knowledge errors. Instead, we introduce a multi-hop Dual Memory Interaction (DMI) mechanism to automatically capture the latent relations among aspect and opinion words. The DMI iteratively infers the correlation vectors of each word by interacting its local memory (LSTM hidden state) with both the global aspect and opinion memories, such that the inter-correlations between aspects and opinions, and the intra-correlations in aspects or opinions can be derived.

Second, how to transfer for this sequence prediction task? One straightforward way is to apply domain adaption methods to align all words within the sentence, however, it is observed that it will not yield significant improvements. Actually, not all the words contribute equally to the domain-invariant feature space though fine-grained adaptation is required. Thus, we propose a novel Selective Adversarial Learning (SAL) method to dynamically learn an alignment weight for each word, where more important words can possess higher alignment weights to achieve a local semantic alignment based on adversarial training. Empirically, the proposed model outperforms the state-of-the-art fine-grained adaptation methods by a large margin on four benchmark datasets. We also conduct extensive ablation studies to quantitatively and qualitatively demonstrate the effectiveness of the selectivity of adversarial learning.

Overall, our main contributions are summarized as: (1) to the best of our knowledge, an unsupervised domain adaptation setting is firstly explored for E2E-ABSA; (2) an effective SAL method is proposed to conduct a local semantic alignment for fine-grained domain adaptation; (3) extensive experiments verify the effectiveness of the proposed SAL method. 

\begin{figure*}[t]
\centering
\includegraphics[width=0.9\textwidth]{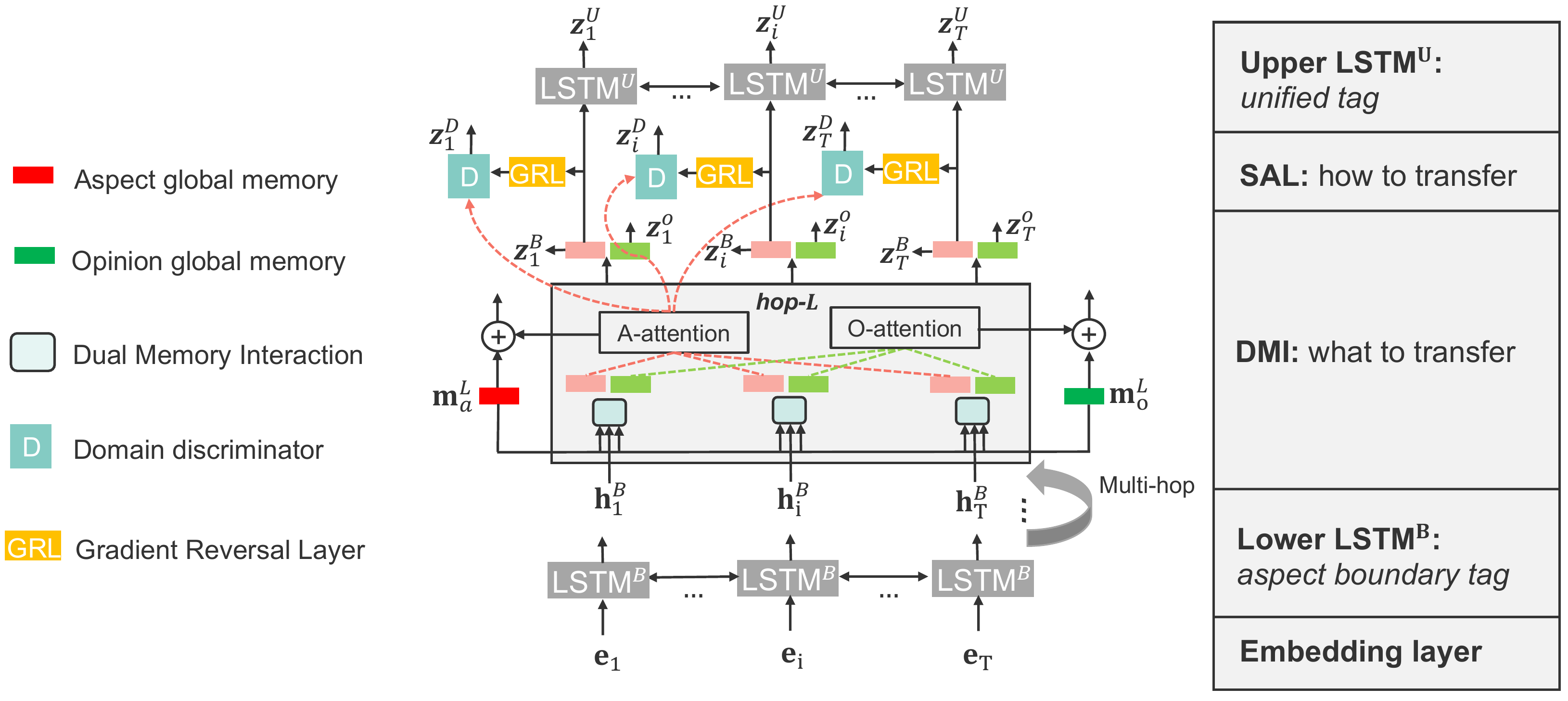}
\vspace{-3mm}
\caption{The framework of the proposed model.}\label{fig:framework}
\end{figure*}

\section{Task Definition}
\label{sec:definition}
\noindent \textbf{Single-domain:} 
E2E-ABSA involves both aspect detection (\textbf{AD}) and aspect sentiment (\textbf{AS}) classification tasks, which are formulated as a unified sequence labeling problem. Given an input sequence of words $\small{\mathbf{x}\!=\!\{w_{1},w_{2},...,w_{T}\}}$ and its word embeddings $\small{\mathbf{e}\!=\!\{e_{1},e_{2},...,e_{T}\}}$, the goal is to predict a tag sequence $\small{\mathbf{y}\!=\!\{y_{1},y_{2},...,y_{T}\}}$ over the {\it unified tags}, with $\small{y_{i} \!\in\! \mathcal{Y}^{\mathcal{U}}\!=\!}$ \{\texttt{B-POS}, \texttt{I-POS}, \texttt{E-POS}, \texttt{S-POS}, \texttt{B-NEG}, \texttt{I-NEG}, \texttt{E-NEG}, \texttt{S-NEG}, \texttt{B-NEU}, \texttt{I-NEU}, \texttt{E-NEU}, \texttt{S-NEU}, \texttt{O}\}. \textbf{Cross-domain:} Here we are performing in a more challenging unsupervised domain adaptation setting. Given a set of labeled data $\small{{D}_{s}\!=\!{\{ (\mathbf{ x }_{ s }^{ i },\mathbf{ y }_{ s }^{ i }) \}  }_{ i=1 }^{ { N }_{ s }}}$ from a source domain and a set of unlabeled data $\small{{D}_{t}\!=\!{\{ (\mathbf{ x }_{ t }^{ j }) \}  }_{ j=1 }^{ { N }_{ t }}}$ from a target domain, we aim to transfer the knowledge of ${D}_{s}$ to improve the sequence learning in ${D}_{t}$.

\section{Model Description}
\noindent \textbf{Overview:} 
As shown in Figure~\ref{fig:framework}, we adopt two stacked bi-directional LSTMs as the base model~\cite{li2019unified} for E2E-ABSA. The upper $\text{LSTM}^{\mathcal{U}}$ is for the {high-level} \textbf{ADS} (AD+AS) task and it predicts the {\it unified tags} as output, while the lower $\text{LSTM}^{\mathcal{B}}$ is for the {low-level} \textbf{AD} task and predicts the {\it aspect boundary tags} as the guidance. To adapt the base model, we design different components in terms of the two problems, i.e., what to transfer and how to transfer, respectively.

(1) To automatically capture the latent relations between aspect and opinion words as transferable knowledge across domains, we introduce a multi-hop Dual Memory Interaction (DMI) mechanism between the two LSTMs. At each hop, e.g., the $1$st hop, each local memory $\mathbf{h}^{\mathcal{B}}_i$ will interact with both the global aspect and opinion memories, i.e., $\small{\mathbf{m}^{1}_{a}}$ and $\small{\mathbf{m}^{1}_{o}}$ based on the DMI, to produce two correlation vectors for aspect and opinion words co-detection, where the opinion detection is used as an auxiliary task for the AD task. The ``local'' memory denotes the hidden representation ($\text{LSTM}^{\mathcal{B}}$ hidden state) of each word within the sentence. Whereas the two ``global'' memories are globally shared by all input sentences, which are commonly used in memory networks~\cite{sukhbaatar2015end,kumar2016ask} and can be seen as high-level representations for aspect and opinion words, respectively. The A-attention and O-attention then aggregate most relevant aspect or opinion words information to refine the two global memories for the next hop. 


(2) To adapt these relations across domains, we propose a Selective Adversarial Learning (SAL) method to dynamically focus on aligning the aspect words between domains. This is because the informative aspect words contribute more to the shared feature space than the unmeaning words tagged with \texttt{O} in the sentence~\cite{joey2019roseq}. As such, an aspect tagger trained on a source domain can work well when applied to a target domain. Specifically, at the final hop, we adopt a domain discriminator for each word with a gradient reversal layer~\cite{ganin2016domain} to perform domain adversarial learning over its correlation vector (alignment). While the A-attention module provides an aspect attention distribution as a selector to control a learnable alignment weight for each word (selectivity). Finally, each aligned correlation vector will be used to predict {\it aspect boundary tags} (AD task) and fed to the $\text{LSTM}^{\mathcal{U}}$ for the {\it unified tags} prediction (ADS task). In the following sections, we detail each component. 

\subsection{Base Model}
We adopt two stacked bi-directional LSTMs as the base model. We link these two LSTM layers so that the hidden representations generated by the $\text{LSTM}^{\mathcal{B}}$ can be fed to $\text{LSTM}^{\mathcal{U}}$ as the guidance information. Specifically, their hidden representations $\mathbf{h}^{\mathcal{B}}_i \in \mathbb{R}^{\text{dim}^{\mathcal{B}}_h}$ and $\mathbf{h}^{\mathcal{U}}_i \in \mathbb{R}^{\text{dim}^{\mathcal{U}}_h}$ at the $i$-th time step ($i \in [1, T]$) are calculated as follows:
\begin{equation*}
\begin{split}
\mathbf{h}^{\mathcal{B}}_i &= [\overrightarrow{\text{LSTM}}^{\mathcal{B}}(\mathbf{e}_i); \overleftarrow{\text{LSTM}}^{\mathcal{B}}(\mathbf{e}_i)],  \\
\mathbf{h}^{\mathcal{U}}_i &= [\overrightarrow{\text{LSTM}}^{\mathcal{U}}(\mathbf{h}^{\mathcal{B}}_i); \overleftarrow{\text{LSTM}}^{\mathcal{U}}(\mathbf{h}^{\mathcal{B}}_i)].\end{split}
\end{equation*}
The probability scores $\small{\mathbf{z}^{\mathcal{B}}_i \!\in\! \mathbb{R}^{|\mathcal{Y}^{\mathcal{B}}|}}$ over the {\it aspect boundary tags} $\mathcal{Y}^{\mathcal{B}}\!=\!$ \{\texttt{B}, \texttt{I}, \texttt{E}, \texttt{S}, \texttt{O}\} are calculated by a fully-connected softmax layer:
\begin{equation*}
\mathbf{z}^{\mathcal{B}}_i \!=\! {\bf p}(\mathbf{y}^{\mathcal{B}}_i|\mathbf{h}^{\mathcal{B}}_i) \!=\! \mathrm{Softmax}(\mathrm{\bf W}_{\mathcal{B}} \mathbf{h}^{\mathcal{B}}_i \!+\! \mathrm{\bf b}_{\mathcal{B}}).
\end{equation*}
Similarly, the scores $\small{\mathbf{z}^{\mathcal{U}}_i \!\in\! \mathbb{R}^{|\mathcal{Y}^{\mathcal{U}}|}}$ over the {\it unified tags} $\mathcal{Y}^{\mathcal{U}}$ defined in Section~\ref{sec:definition} are obtained as:
\begin{equation*}
\mathbf{z}^{\mathcal{U}}_i \!=\! {\bf p}(\mathbf{y}^{\mathcal{U}}_i |\mathbf{h}^{\mathcal{U}}_i) \!=\! \mathrm{Softmax}(\mathrm{\bf W}_{\mathcal{U}} \mathbf{h}^{\mathcal{U}}_i \!+\! \mathrm{\bf b}_{\mathcal{U}}).
\end{equation*}

\subsection{Global-Local Memory Interaction}
\label{sec:GLMI}
Before detailing the DMI module, we firstly introduce Global-Local Memory Interaction (GLMI) that describes
the interaction between a local memory $\small{\mathbf{h}^{}_{i}\!\in\! \mathbb{R}^{\text{dim}_h}}$ and a global memory $\small{\mathbf{m} \!\in\! \mathbb{R}^{\text{dim}_h}}$. 
Formally, we parameterize the GLMI  $\small{f(\mathbf{h}_{i}, \mathbf{m}; \bm{\Theta }, \mathbf{G})}$, with $\small{\bm{\Theta }\!=\!\{\mathbf{W}, \mathbf{b}\}}$ and $\small{\mathbf{G}}$, which consists of a residual transformation and a tensor product operation. Specifically, we firstly incorporate the global memory information $\mathbf{m}$ into each local position with a residual transformation as $\small{\tilde { \mathbf{h}}_i \!=\!\mathbf{h}_i  + \mathrm{ReLU} (\mathbf{W} [\mathbf{h_i}\!:\!\mathbf{m}] +{\mathbf b})}$, where $[:]$ denotes the vector concatenation. As such, the global memory can distill more correlated local information and they are mapped into the same space. Then we compute a correlation vector $\mathbf{r}_{i}\!\in\! \mathbb{R}^{K}$ that encodes the strength of correlations between the global memory $\mathbf{m}$ and the transformed local memory $\tilde { \mathbf{h}}_i$  through a  tensor product operation as:
\begin{equation*}
\mathbf{r}_{i} = {\mathbf{m}}^{T}  \mathbf{G} \tilde { \mathbf{h}}^{\mathcal{B}}_i ,
\end{equation*}
where the tensor $\small{\mathbf{G} \!\in\! \mathbb{R}^{ \text{dim}_h \times \text{dim}_h \times K}}$ can be seen as multiple bilinear matrice that model $K$ kinds of latent relations between two objects. The $k$-th slice of the $\mathbf{G}$, i.e., $\small{\mathbf{G}_{k}\!\in\! \mathbb{R}^{\text{dim}_h \times \text{dim}_h}}$ denotes one type of latent relation that interacts with 2 vectors to constitute one type of composition.

\begin{figure}[t]
\centering
\includegraphics[width=0.8\columnwidth]{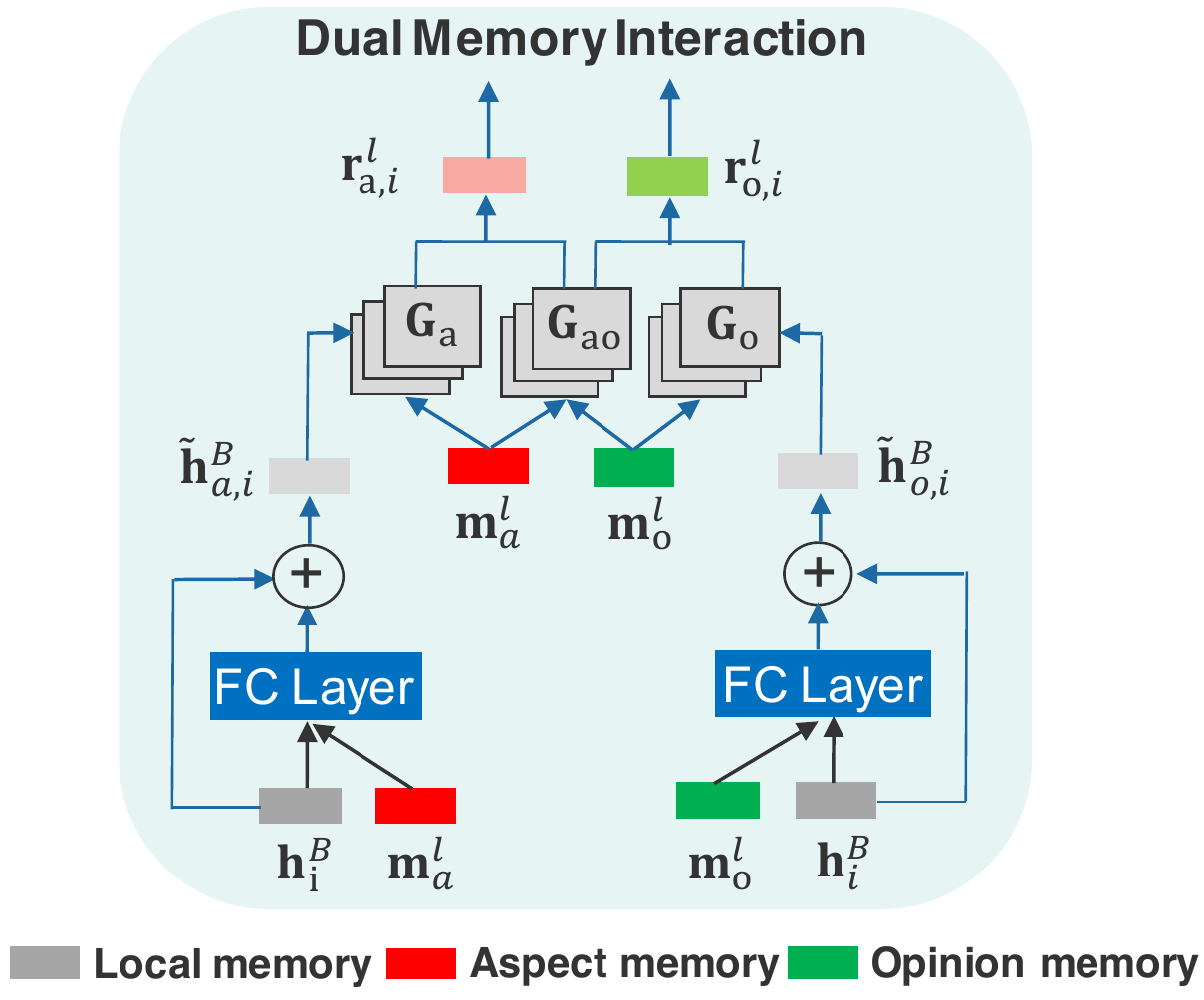}
\caption{The Dual Memory Interaction (DMI).}\label{fig:du_inter}
\vspace{-3mm}
\end{figure}

\subsection{Dual Memory Interaction}
Following the notations in Section~\ref{sec:GLMI}, we further define a global aspect memory $\small{\mathbf{m}_{a} \!\in\! \mathbb{R}^{\text{dim}^{\mathcal{B}}_h}}$, a global opinion memory $\small{\mathbf{m}_{o} \!\in\! \mathbb{R}^{\text{dim}^{\mathcal{B}}_h}}$, and $\text{LSTM}^{\mathcal{B}}$ hidden states $\small{\mathbf{H}^{\mathcal{B}}\!=\!{ \{ \mathbf{h}^{\mathcal{B}}_{ i } \}  }_{ i=1 }^{ T } }$ as the local memories. The global aspect and opinion memories are able to capture highly correlated aspect or opinion words from the local memories, respectively. Based on the observation that aspect words are often collocated with opinion words across domains, thus their associations can act as the pivot information to bridge the domain gaps. To automatically capture their latent relations within the sentences, at the $l$-th hop, each local memory ${ \mathbf{h}}^{\mathcal{B}}_i $ will interact with the global memories $\mathbf{m}^{l}_{a}$ and $\mathbf{m}^{l}_{o}$ by the Dual Memory Interaction (DMI) shown in Figure~\ref{fig:du_inter}, to produce two correlation vectors for aspect and opinion co-detection:
\begin{equation*}  
\begin{split}
\mathbf{r}^{l}_{a,i} &\!=\! [f(\mathbf{h}^{\mathcal{B}}_{i}, \mathbf{m}^{l}_{a}; \bm{\Theta }_{a}, \mathbf{G}_{a})\!:\!f(\mathbf{h}^{\mathcal{B}}_{i}, \mathbf{m}^{l}_{o}; \bm{\Theta }_{o}, \mathbf{G}_{ao})],\\
\mathbf{r}^{l}_{o,i} &\!=\! [f(\mathbf{h}^{\mathcal{B}}_{i}, \mathbf{m}^{l}_{o}; \bm{\Theta }_{o}, \mathbf{G}_{o})\!:\!f(\mathbf{h}^{\mathcal{B}}_{i}, \mathbf{m}^{l}_{a}; \bm{\Theta }_{a}, \mathbf{G}^\top_{ao})],
\end{split}
\end{equation*}
where $\small{\mathbf{G}_{a},\mathbf{G}_{o}}$ and  $\small{\mathbf{G}_{ao}}$ denote the composition tensors of modeling the latent relations between aspect and aspect, opinion and opinion, and aspect and opinion, respectively.

The correlation vector measures the association strength between local and global memories; e.g., If ${ \mathbf{h}}^{\mathcal{B}}_i $ for the word ${{w}_{i}}$ is both highly intra-correlated with the aspect memory $\small{\mathbf{m}_{a}} $ and inter-correlated with the opinion memory $\small{\mathbf{m}_{o}}$, ${{w}_{i}}$ is more likely to be an aspect term. Then the two correlation vectors can be transformed to a scalar \textbf{a}spect attention (\textbf{A}-attention) and \textbf{o}pinion attention (\textbf{O}-attention) weight ${\alpha}^{l}_{p,i} $, respectively, with $\small{p \!\in\! \{a, o\}}$ denoting the aspect or opinion, which indicates the possibility of each word in the sentence being an aspect word or an opinion word as:
\begin{equation*}  
{\alpha}^{l}_{p,i} = \frac { {\exp}(\mathbf{W}_{p} \mathbf{r}^{l}_{p,i}) }{ \sum _{ j=1 }^{ T }{ {\exp}(\mathbf{W}_{p} \mathbf{r}^{l}_{p,j}) }  },
\end{equation*}
where $\mathbf{W}_{p}$ is the weight of the attention module. The aspect or opinion attention weight ${\alpha}^{l}_{p,i}$ will summarize the local memories to update the global aspect and opinion memories, respectively, for the next hop, i.e., $\small{\mathbf{m}^{l+1}_{p} \!=\! \mathbf{m}^{l}_{p} + \sum _{ i=1 }^{ T }{{\alpha}^{l}_{p,i} \mathbf{h}^{\mathcal{B}}_{i}}}$.
The updates gradually refine the global memories to incorporate more relevant candidates based on the attention mechanism. In the DMI, all parameters are shared in different hops and domains.

At the final $L$-th hop, we use $\mathbf{r}^{L}_{a,i}$ for the AD task and feed it to the $\text{LSTM}^{\mathcal{U}}$ for the ADS task. For the auxiliary opinion detection task, we feed $\mathbf{r}^{L}_{o,i}$ into a softmax layer for predicting the probability scores $\small{\mathbf{z}^{\mathcal{O}}_i \!\in\! \mathbb{R}^{|\mathcal{Y}^{\mathcal{O}}|}}$ over the opinion labels\footnote{The opinion lexicon  (\url{http://mpqa.cs.pitt.edu/}) is used to provide opinion labels for both domains.} $\mathcal{Y}^{\mathcal{O}}$, i.e., a word is an opinion word or not, as:
\begin{equation*}
    \mathbf{z}^{\mathcal{O}}_i \!=\! {\bf p}(\mathbf{y}^{\mathcal{O}}_i|\mathbf{r}^{L}_{o,i}) \!=\! \mathrm{Softmax}(\mathrm{\bf W}_{\mathcal{O}}\mathbf{r}^{L}_{o,i}  \!+\! \mathrm{\bf b}_{\mathcal{O}}).
\end{equation*}

\subsection{Selective Adversarial Learning}
To adapt the captured relations to be domain-invariant, we propose a Selective Adversarial Learning (SAL) method to dynamically align the words with high probability to fall into the aspect boundaries, i.e., being an aspect word. Specifically, we introduce a domain discriminator for each word, which aims to identify the domain label $\small{\mathbf{y}^{\mathcal{D}}_i \!\in\! \mathbb{R}^{|\mathcal{Y}^{\mathcal{D}}|}}$ of the input word, i.e., the word in the sentence is from the source or the target domain. While the feature extractor is to produce the domain-invariant correlation vector $\mathbf{r}^{L}_{a,i}$ that cannot be distinguished by the domain discriminator via a Gradient Reversal Layer (GRL)~\cite{ganin2016domain}. Mathematically, we formulate the GRL as a `pseudo-function' $\small{{ R }_{ \lambda  }(\mathbf{x})\!=\!\mathbf{x}}$ with a reversal gradient $\small{\frac { \partial { R }_{ \lambda  }(\mathbf{x}) }{ \partial \mathbf{x} } \!=\!-\lambda I}$, where $\lambda$ is the adaptation rate. The correlation vector $\mathbf{r}^{L}_{a,i}$ will be fed to the GRL before the domain discriminator, which is used to predict the probability scores $\small{\mathbf{z}^{\mathcal{D}}_i \in \mathbb{R}^{|\mathcal{Y}^{\mathcal{D}}\mathbf{|}}}$ over the domain labels $\mathcal{Y}^{\mathcal{D}}$ as:
\begin{equation*}
\mathbf{z}^{\mathcal{D}}_i \!=\! {\bf p}(\mathbf{y}^{\mathcal{D}}_i|\mathbf{r}^{L}_{a,i}) \!=\! \mathrm{Softmax}(\mathrm{\bf W}_{\mathcal{D}}{ R }_{ \lambda  }(\mathbf{r}^{L}_{a,i})  \!+\! \mathrm{\bf b}_{\mathcal{D}}).
\end{equation*}
And meanwhile, the aspect attention weight ${\alpha}^{L}_{a,i}$ at the final hop serves as a selector to be a learnable alignment weight for each word. Thus, the selective domain adversarial loss is a weighted cross-entropy loss $\ell$ for all the words from the labeled source data $D_{s}$ and unlabeled target data $D_{t}$:
\begin{equation}  
\label{eqn:domain_loss}
\mathcal{L}_{\mathcal{D}}=\sum _{ {D}_{s}\cup{D}_{t}   }\sum _{i=1}^{T}{\alpha}^{L}_{a,i} \ell (\mathbf{z}^{\mathcal{D}}_i, \mathbf{y}^{\mathcal{D}}_i).
\end{equation}
Existing studies~\cite{yosinski2014transferable,mou2016transferable} have already shown some evidence that low-level neural layer features (i.e., low-level task) are more easily transferred to different tasks or domains. Thus, we choose the $\mathbf{r}^{L}_{a,i}$ from the low-level AD task to be aligned instead of the feature $\mathbf{h}^{\mathcal{U}}_{i}$ from the high-level ADS task to transfer. Our ablation studies also confirm this assumption.

\subsection{Alternating Training}
The primary task loss consists of the cross-entropy losses $\ell$ for both the guided AD and main ADS tasks for the labeled source data $D_{s}$:
\begin{equation} 
\label{eqn:main_loss}
\mathcal{L}_{\mathcal{M}}=\sum _{ {D}_{s} }\sum _{ \mathcal{Q} \in \{\mathcal{B}, \mathcal{U}\}  }\sum _{i=1}^{T}\ell (\mathbf{z}^{\mathcal{Q}}_i, \mathbf{y}^{\mathcal{Q}}_i).
\end{equation}
The auxiliary opinion detection loss is the cross-entropy loss for the labeled source data $D_{s}$ and unlabeled target data $D_{t}$ as follows:
\begin{equation}  
\label{eqn:aux_loss}
\mathcal{L}_{\mathcal{O}}=\sum _{ {D}_{s}\cup{D}_{t}   }\sum _{i=1}^{T}\ell (\mathbf{z}^{\mathcal{O}}_i, \mathbf{y}^{\mathcal{O}}_i).
\end{equation}
Traditionally, we can directly optimize the joint loss of Eqs. (\ref{eqn:domain_loss})-(\ref{eqn:aux_loss}), i.e., $\small{E\!=\!\mathcal{L}_{\mathcal{M}}\!+\!\rho\mathcal{L}_{\mathcal{O}}\!+\!\gamma \mathcal{L}_{\mathcal{D}}}$ to obtain both discriminative and domain-invariant word representations, where $\rho$ and $\gamma $ are the trade-off factors. However, we found the optimization process tends to be unstable since it may be hard to jointly optimize many objectives. Thus, we propose an empirically alternating strategy to train the $\small{\mathcal{L}_{\mathcal{M}}\!+\!\rho\mathcal{L}_{\mathcal{O}}}$ and $\small{\mathcal{L}_{\mathcal{D}}}$ iteratively, which separates the whole word representation learning into a {\it discriminative} stage and a {\it domain-invariant} stage. Let $\bm{\theta}_{f}$, $\bm{\theta}_{w}$, $\bm{\theta}_{d}$ denote the parameters for feature learning of each word, word predictors for AD, ADS and opinion detection tasks, and domain discriminators, respectively. Based on our strategy, we are seeking the parameters ($\hat{\bm{\theta}}_{f}^{(1)}$, $\hat{\bm{\theta}}_{f}^{(2)}$, $\hat{\bm{\theta}}_{w}$, $\hat{\bm{\theta}}_{d}$) that deliver a saddle point of $E$ among two stages:
\begin{equation*} 
\begin{split}
(\hat{\bm{\theta}}_{f}^{(1)}, \hat{\bm{\theta}}_{w})&\!=\! \arg \underset {{\bm{\theta}}_{f}, {\bm{\theta}}_{w}  }{ \min } \mathcal{L}_{\mathcal{M}}+\rho\mathcal{L}_{\mathcal{O}}\\
(\hat{\bm{\theta}}_{f}^{(2)}, \hat{\bm{\theta}}_{d})&\!= \! \arg \underset { {\bm{\theta}}_{d} }{ \min }\ \underset { {\bm{\theta}}_{f}}{ \max }\  \mathcal{L}_{\mathcal{D}}.
\end{split}
\end{equation*}
At the saddle point, the feature learning parameters $\bm{\theta}_{f}$ minimize the word prediction losses (i.e., the features are discriminative) for the first stage. For the second stage, the domain classification loss is minimized by the domain discriminator parameters $\bm{\theta}_{d}$ while maximized by the feature learning parameters $\bm{\theta}_{f}$ via GRL (i.e., the features are domain-invariant). As such, we can achieve easier and more stable optimization for feature learning.

\section{Experiments}
\subsection{Experimental Setup}
\noindent \textbf{Datasets:} 
Our experiments are conducted on four benchmark datasets: Laptop (${\mathbb{L}}$), Restaurant ($\mathbb{R}$), Device ($\mathbb{D}$), and Service ($\mathbb{S}$). $\mathbb{L}$ contains reviews from the laptop domain in SemEval ABSA challenge 2014~\cite{pontiki2014semeval}. Following the setup in \cite{li2019unified}, $\mathbb{R}$ is the union set of the restaurant datasets from SemEval ABSA challenge 2014, 2015, and 2016~\cite{pontiki2014semeval,pontiki2015semeval,pontiki2016semeval}. $\mathbb{D}$ is a combination of device reviews from 5 different digital products provided by~\cite{hu2004mining}. $\mathbb{S}$ is introduced by~\cite{toprak2010sentence} and contains reviews from web services. Detailed statistics are shown in Table~\ref{tab:dataset}. 

\begin{table}[t]\small
    \centering
    \begin{tabular}{c|c|c|c|c}
    \Xhline{3\arrayrulewidth}
      Dataset &Domain &Sentences &Training &Testing  \\ \hline
        $\mathbb{L}$    & Laptop  & 1,869  & 1,458 & 411  \\ \hline
        $\mathbb{R}$   & Restaurant  & 3,900  & 2,481 & 1,419  \\ \hline
        $\mathbb{D}$   & Device  & 1,437  & 954 & 483  \\ \hline
        $\mathbb{S}$   & Service  & 2,153  & 1,433 & 720  \\ 
       \Xhline{3\arrayrulewidth}
    \end{tabular}
    \caption{Statistics of the datasets.}
    \vspace{-6mm}
    \label{tab:dataset}
\end{table}

\noindent \textbf{Settings:} 
We construct 10 transfer pairs like ${D}_{s}$$\rightarrow$${D}_{t}$ with the four domains mentioned above, and we do not use the pairs $\mathbb{L}$$\rightarrow$$\mathbb{D}$ and $\mathbb{D}$$\rightarrow$$\mathbb{L}$ as these two domains are very similar. Note that for the unsupervised domain adaptation setting, no labels are available for the target domain. Therefore, for each transfer pair, its training dataset is the combination of the labeled training data of the source domain and the unlabeled training data of the target domain. Meanwhile, it employs the testing data of the source domain with labels as the validation set and the testing data of the target domain as the evaluation set. We report the results for both AD and ADS tasks.
The evaluation metric is the Micro-F1 score under the \textbf{exact} match, which means that an output segment is considered to be correct only if it exactly matches with the gold standard span of the aspect term for the AD task or the aspect term and its sentiment for the ADS task. All experiments are repeated 5 times and we report the average results over 5 runs. 

\begin{table*}[htb] 
\centering
\resizebox{1.8\columnwidth}{!}
{\begin{tabular}{c|c|c|c|c|c|c|c|c|c|c|c|c|c|c}
    \Xhline{3\arrayrulewidth}
 \multirow{2}{*}{Transfer Pair} &  \multicolumn{2}{c|}{{TCRF}}  & \multicolumn{2}{c|}{RAP}  &
 \multicolumn{2}{c|}{Hier-Joint} &\multicolumn{2}{c|}{$\text{Hier-Joint}^{+}$} & \multicolumn{2}{c|}{RNSCN}  & \multicolumn{2}{c|}{$\text{RNSCN}^{+}$} & \multicolumn{2}{c}{\textbf{Ours}}  \\
\cline{2-15}    & AD & ADS & AD & ADS & AD & ADS  & AD & ADS & AD & ADS & AD & ADS & AD & ADS\\ \hline \hline
$\mathbb{S}$$\rightarrow$$\mathbb{R}$      & - & 14.84  &- & 25.41 &- & 32.81 & 46.39 & 31.10 &- & 30.56 & {48.89} & 33.21 &\textbf{52.05} &\textbf{41.03}                 \\
$\mathbb{L}$$\rightarrow$$\mathbb{R}$      & - & 16.06  &- & 31.05 &- & 31.90 & 48.61 & 33.54 &- & 31.85 & 52.19 & 35.65 &\textbf{56.12} &\textbf{43.04}               \\ 
$\mathbb{D}$$\rightarrow$$\mathbb{R}$      & - & 17.05 &- & 28.37 &- & 30.03 & 42.96 & 32.87 &- & 31.41 & 50.39 & 34.60 &\textbf{51.55} &\textbf{41.01}                \\\hline
$\mathbb{R}$$\rightarrow$$\mathbb{S}$      & - & 15.20 &- & 13.17 &- & 15.20 & 27.18 & 15.56 &- & 23.31 & 30.41 & 20.04 &\textbf{39.02} &\textbf{28.01}               \\
$\mathbb{L}$$\rightarrow$$\mathbb{S}$      & - & 12.34  &- & 13.72 &- & 15.33 & 25.22 & 13.90 &- & 16.73 & 31.21 & 16.59 &\textbf{38.26} &\textbf{27.20}               \\
$\mathbb{D}$$\rightarrow$$\mathbb{S}$      & - & 13.49 &- & 16.80 &- & 18.74 & 29.28 & 19.04 &- & 18.93 & {35.50} & 20.03 &\textbf{36.11}   &\textbf{26.62}              \\ \hline
$\mathbb{R}$$\rightarrow$$\mathbb{L}$      & - & 14.59 &- & 15.69 &- & 19.17 & 34.11 & 20.72 &- & 25.54 &\textbf{47.23} & 26.63 &{45.01} &\textbf{34.13}                \\
$\mathbb{S}$$\rightarrow$$\mathbb{L}$      & - & 9.56  &- & 12.38 &- & 21.80 & 33.02 & {22.65} &- & 19.15 &{34.03} & 18.87 &\textbf{35.99} &\textbf{27.04}               \\\hline
$\mathbb{R}$$\rightarrow$$\mathbb{D}$      & - & 19.84  &- & 17.50 &- & 22.91 & 34.81 & 24.53 &- & 32.43 &\textbf{46.16} & 33.26 &{43.76} &\textbf{35.44}              \\
$\mathbb{S}$$\rightarrow$$\mathbb{D}$      & - & 13.43  &- & 15.74 &- & 20.04 &{35.00} & 23.24 &- & 19.98 &32.41 & 22.00 &\textbf{41.21} &\textbf{33.56}              \\ \hline
\hline
Average &- &14.64    &- & 18.98 &- & 22.79 & 35.66 & 23.72 &- & 24.99 &40.84 &26.09  &\textbf{43.91}$^{\dag}$	&\textbf{33.71}$^{\dag}$              \\ 
($\Delta$) &- &(19.07)     &- & (14.73) &- & (10.92) &(8.25) &(9.99) &- & (8.72)  &(3.07)	&(7.62)        &  - & -      \\ 
        \Xhline{3\arrayrulewidth}
\end{tabular}}
\caption{Main results (\%). $\Delta$ refers to the improvements of the full model over baseline methods. The marker $^{\dag}$ means that our model significantly outperforms the best baseline $\textbf{RNSCN}^{+}$ with $p$-value $<$ 0.01.}
\label{tab:main_results}
\vspace{-2mm}
\end{table*}

\subsection{Implementation details}
The word embeddings are $100$-dimensional \emph{word2vec}~\cite{mikolov2013distributed} vectors pre-trained on the combination of the Yelp Challenge dataset\footnote{\url{http://www.yelp.com/dataset challenge}} and the electronics dataset from Amazon reviews\footnote{\url{http://jmcauley.ucsd.edu/data/amazon/links.html}}. For out-of-vocabulary words, we randomly initialized them with a uniform distribution $\small{U( -0.25,0.25 )}$. The dimensions of two LSTM layers ${\text{dim}^{\mathcal{B}}_h}$ and ${\text{dim}^{\mathcal{U}}_h}$ are all set to 100. The number of hops $L$ is set to 2. The number of bilinear interactions $K$ is set to 50. The weight matrices are initialized with a uniform distribution $\small{U( -0.2,0.2 )}$. The adaptation rate $\lambda$ is 0.1 and the trade-off factor $\rho$ is 1.0. For training, the model is optimized by the Adam algorithm \cite{kingma2014adam} with the initial learning rate 0.001. The batch size is 64, with a half coming from the source and target domains, respectively. Gradients with the $\ell_2$ norm larger than 40 are normalized to be 40. To alleviate the overfitting, we apply the dropout on the word embeddings $\mathbf{e}_{i}$ and the learned word representations $\mathbf{r}^{l}_{a,i}$, $\mathbf{r}^{l}_{o,i}$, and $\mathbf{h}^{\mathcal{U}}_i $ with dropout rate 0.5. We use the same hyper-parameters, which are tuned on 10\% randomly held-out training data of the source domain in $\mathbb{R}$$\rightarrow$$\mathbb{L}$, for all transfer pairs.


\subsection{Baselines} 
We compare with several state-of-the-art fine-grained adaptation methods.

\begin{itemize}
    \vspace{-1mm}
    \item \textbf{TCRF}~\cite{jakob2010extracting}: Transferable CRF that uses a linear-chain CRF for sequence prediction based on shared non-lexical features across domains, e.g., POS tags and dependency relations.
    \vspace{-1mm}
    \item \textbf{RAP}~\cite{li2012cross}: A cross-domain Relational Adaptive Bootstrapping method that iteratively expands target aspect and opinion lexicons according to common opinion words and syntactic relations.
    \vspace{-1mm}
    \item \textbf{Hier-Joint}~\cite{ding2017recurrent}: A recurrent neural network (RNN) with manually designed rule-based auxiliary tasks based on common syntactic relations among aspect and opinion words.
    \vspace{-1mm}
    \item \mbox{\textbf{RNSCN}}~\cite{wang2018recursive}: a recursive neural structural correspondence network that incorporates syntactic structures and exploits an auto-encoder to denoise relation labels generated from the parser.
\end{itemize}

\begin{table*}[htb] \small
\centering
\resizebox{2.0\columnwidth}{!}
{
    \begin{tabular}{c|c|c|c|c|c|c|c|c|c|c|c|c}
        \Xhline{3\arrayrulewidth}
 \multirow{3}{*}{Transfer Pair} & \multicolumn{2}{c|}{\textbf{Lower bound}} &  \multicolumn{6}{c|}{\textbf{Ablation Models}}  & \multicolumn{2}{c|}{\textbf{Full Model}}  & \multicolumn{2}{c}{\textbf{Upper bound}}  \\          \cline{2-13} 
  & \multicolumn{2}{c|}{Base Model (SO)} &  \multicolumn{2}{c|}{Base Model+{DMI}}  & \multicolumn{2}{c|}{{AD-AL}}  & \multicolumn{2}{c|}{{ADS-SAL}} & \multicolumn{2}{c|}{{AD-SAL}} & \multicolumn{2}{c}{{Base Model (TO)}} \\ 
\cline{2-13}    & AD & ADS & AD & ADS & AD & ADS & AD & ADS & AD & ADS & AD & ADS \\ \hline  \hline
$\mathbb{S}$$\rightarrow$$\mathbb{R}$     &30.32    &19.74      &45.68   &37.10   &48.28   &37.65  &51.29   &41.03  &\textbf{52.05}   &\textbf{41.03}    & \multirow{3}{*}{81.84} & \multirow{3}{*}{67.26}       \\
$\mathbb{L}$$\rightarrow$$\mathbb{R}$     &33.99    &28.34      &46.25   &36.49   &51.79   &38.63  &55.50   &42.00  &\textbf{56.12}   &\textbf{43.04}    & &     \\ 
$\mathbb{D}$$\rightarrow$$\mathbb{R}$     &31.59    &27.25      &46.56   &36.89   &46.39   &37.34  &{46.43}   &38.35  &\textbf{51.55}   &\textbf{41.01}   & &      \\\hline
$\mathbb{R}$$\rightarrow$$\mathbb{S}$     &15.63    &8.61       &21.88   &16.85   &25.13   &18.61  &37.11   &25.84  &\textbf{39.02}   &\textbf{28.01}   & \multirow{3}{*}{68.28} & \multirow{3}{*}{41.12}      \\
$\mathbb{L}$$\rightarrow$$\mathbb{S}$     &22.45    &16.07      &28.67   &21.53   &28.18   &20.74  &30.35   &23.73  &\textbf{38.26}   &\textbf{27.20}   & &      \\
$\mathbb{D}$$\rightarrow$$\mathbb{S}$     &16.79    &9.49       &31.91   &22.14   &32.88   &24.89  &32.51   &21.45  &\textbf{36.11}   &\textbf{26.62}   & &     \\ \hline
$\mathbb{R}$$\rightarrow$$\mathbb{L}$     &38.45    &23.40      &42.27   &30.52   &40.52   &28.77  &44.56   &{33.34}  &\textbf{45.01}   &\textbf{34.13}    & \multirow{2}{*}{75.95} & \multirow{2}{*}{52.62}      \\
$\mathbb{S}$$\rightarrow$$\mathbb{L}$     &24.69    &14.48      &36.38   &27.48   &32.96   &25.16  &33.87   &{24.22}  &\textbf{35.99}   &\textbf{27.04}    & &     \\\hline
$\mathbb{R}$$\rightarrow$$\mathbb{D}$     &34.87    &25.79      &36.90   &27.71   &41.61   &31.88  &\textbf{43.97}   &34.50  &{43.76}   &\textbf{35.44}  & \multirow{2}{*}{70.37} & \multirow{2}{*}{57.62}      \\
$\mathbb{S}$$\rightarrow$$\mathbb{D}$     &27.73    &17.73      &38.03   &31.21   &39.54   &32.28  &{40.40}   &{33.26}  &\textbf{41.21}   &\textbf{33.56}  & &      \\ \hline \hline
Average  &27.65	&19.09	&37.45	&28.79	&38.73	&29.60	&41.60	&31.77	&\textbf{43.91}$^{\dag}$	&\textbf{33.71}$^{\dag}$ & 74.11 & 54.66            \\ 
($\Delta$)  &(16.26)	&(14.62)	&(6.46)	&(4.92)	&(5.18)	&(4.11)	&(2.31)	&(1.94)	& -	&  - & -	&  -           \\ 
        \Xhline{3\arrayrulewidth}
\end{tabular}}
\caption{Ablation results (\%). $\Delta$ refers to the improvements of the full model over ablation methods. The marker $^{\dag}$ means that the full model significantly outperforms the best ablation model \textbf{ADS-SAL} with $p$-value $<$ 0.01.}
\label{tab:ablation_results}
\vspace{-2mm}
\end{table*}

As the first to address cross-domain E2E-ABSA, we have to adapt all the baselines which are originally proposed for cross-domain aspect detection, or aspect and opinion co-detection to return the ADS results by replacing their {\it aspect boundary tags} with the {\it unified tags}.
Absent of the proposed stacking architecture, all the baselines fail to accomplish the auxiliary AD task meantime. Thus, we only report their ADS results.
Besides, E2E-ABSA aims to simultaneously learn aspect terms along with their sentiments. Thus, the ADS is exactly our main task while the AD is only an auxiliary task used for the guidance.

To be more convincing, we extend neural models (i.e., {Hier-Joint} and {RNSCN}) to more powerful baselines named $\textbf{Hier-Joint}^{+}$ and $\textbf{RNSCN}^{+}$ with the proposed stacking architecture, respectively. Both of them stack an additional RNN layer on top of the original framework to produce the {\it unified tags} while the lower layer is to predict the {\it aspect boundary tags} as the guidance. The validity of such extensions is guaranteed by the fact that the extended versions achieve even better AD performances than the original versions. We use the source code of the baselines for experiments. For fair comparison, all baselines use the same pre-trained word embeddings and the baselines that require opinion labels use the same opinion lexicon.

\subsection{Main Results}
Based on the results in Table~\ref{tab:main_results}, we have the following observations:




\begin{itemize}
    \vspace{-1mm}
    \item Our model consistently and significantly achieves the best results on almost all transfer pairs, outperforming the strongest baseline $\text{RNSCN}^{+}$ by 3.07\% and 7.62\% Micro-F1 on average for the AD and ADS tasks, respectively. Our model can automatically model complicated relations among aspect and opinion words via the DMI as transferable knowledge. Besides, the proposed SAL method can dynamically learn an alignment weight for each word to achieve a local semantic alignment, which distills a better shared feature space and further improves the performances.
    \vspace{-1mm}
    \item Traditional non-neural methods like TCRF and RAP perform very poorly due to the reliance on hand-crafted features. Our model outperforms Hier-Joint and RNSCN, which are neural models, by 10.92\% and 8.72\% Micro-F1 on average for the ADS task, respectively. Both of them rely on the dependency parser to exploit syntactic relations, which are inflexible due to the no end-to-end manner and may bring in external errors.
    \vspace{-1mm}
    \item The extended version $\text{Hier-Joint}^{+}$ and $\text{RNSCN}^{+}$ can further improve the performances, which shows the benefits of the guidance from the low-level AD task. However, our model can still outperform them by a large margin, which demonstrates the effectiveness of the proposed methods.
\end{itemize}

\subsection{Ablation Study}
To investigate the effectiveness of each component, we conduct the ablation study to compare our full model with different ablation variants: 
\begin{itemize}
\vspace{-1mm}
\item \textbf{Base Model (SO / TO)}: it uses two stacked Bi-LSTMs as the Base Model. {SO} (Source Only) and {TO} (Target Only) denote that the base model is only trained on the labeled data from the source and target domain, respectively. We usually refer to them as a lower bound and a upper bound, respectively.
\vspace{-1mm}
\item \textbf{Base Model+DMI}: it uses two stacked Bi-LSTMs with a multi-hop dual memory interaction ({DMI}) between them.
\vspace{-1mm}
\item \textbf{AD-AL}: it performs pure adversarial learning (removing the selective weight ${\alpha}^{L}_{a,i}$ from the Eq. (\ref{eqn:domain_loss})) on each correlation vector $\mathbf{r}^{L}_{a,i}$ for the low-level AD task.
\vspace{-1mm}
\item \textbf{AD-SAL}: it advances the AD-AL by conducting selective adversarial learning.
\vspace{-1mm}
\item \textbf{ADS-SAL}: it conducts selective adversarial learning on each word representation $\mathbf{h}^{\mathcal{U}}_{i}$ for the high-level ADS task.
\vspace{-1mm}
\end{itemize}

\begin{table*}[t]
    \centering
    \resizebox{2.1\columnwidth}{!}
    {%
    \begin{tabular}{@{}L{6.9cm}@{~}|@{~}L{1.8cm}@{~}|@{~}L{2.6cm}@{~}|@{~}L{1.8cm}@{~}|@{~}L{2.6cm}@{~}|@{~}L{3.0cm}@{~}|@{~}L{3.7cm}@{}}
    \Xhline{3\arrayrulewidth}
        \multirow{2}{*}{\textbf{Input: (Target domain $\mathbb{L}$)}} & \multicolumn{2}{c|@{~}}{Base model+DMI} & \multicolumn{2}{c|@{~}}{AD-AL} & \multicolumn{2}{c}{{AD-SAL}} \\ \cline{2-7}
        & AD & ADS & AD & ADS & AD & ADS  \\ \hline
        1. This laptop has only 2 \textbf{[}\textcolor{red}{\textit{usb ports}}\textbf{]}$_{\texttt{NEG}}$ , and they are both on the same side . & \textit{ports}(${\text{\xmark}}$), \textit{side} (${\text{\xmark}}$) & NONE(${\text{\xmark}}$) &   NONE(${\text{\xmark}}$)  & NONE(${\text{\xmark}}$)  & \textit{usb ports} & [\textit{usb ports}]$_{\texttt{NEG}}$ \\ \hline
        2. It is very easy to integrate \textbf{[}\textcolor{red}{\textit{bluetooth devices}}\textbf{]}$_{\texttt{POS}}$ , and \textbf{[}\textcolor{red}{\textit{usb devices}}\textbf{]}$_{\texttt{POS}}$ are recognized almost instantly . & \textit{devices} (${\text{\xmark}}$), \textit{devices} (${\text{\xmark}}$)  & [\textit{devices}]$_{\texttt{POS}}$ (${\text{\xmark}}$), [\textit{devices}]$_{\texttt{POS}}$ (${\text{\xmark}}$) &   NONE(${\text{\xmark}}$)  & NONE(${\text{\xmark}}$)  & \textit{bluetooth devices}, \textit{usb devices} & [\textit{bluetooth devices}]$_{\texttt{POS}}$,  [\textit{usb devices}]$_{\texttt{POS}}$ \\ \hline
        3. I also wanted \textbf{[}\textcolor{red}{\textit{windows 7}}\textbf{]}$_{\texttt{POS}}$ , which this one has . & NONE(${\text{\xmark}}$)  & NONE(${\text{\xmark}}$) &   NONE(${\text{\xmark}}$)  & NONE(${\text{\xmark}}$)  & \textit{windows 7} & [\textit{windows 7}]$_{\texttt{POS}}$ \\ \hline
        4. The \textbf{[}\textcolor{red}{\textit{speed}}\textbf{]}$_{\texttt{POS}}$ , the \textbf{[}\textcolor{red}{\textit{simplicity}}\textbf{]}$_{\texttt{POS}}$ , the \textbf{[}\textcolor{red}{\textit{design}}\textbf{]}$_{\texttt{POS}}$ it is lightyears ahead of any pc i have ever owned . &  \textit{speed}, \textit{design}    & [\textit{speed}]$_{\texttt{POS}}$, [\textit{design}]$_{\texttt{POS}}$ &   \textit{speed}, \textit{design}, \textit{pc} (${\text{\xmark}}$)  &  [\textit{speed}]$_{\texttt{POS}}$, [\textit{design}]$_{\texttt{POS}}$, [\textit{pc}]$_{\texttt{POS}}$ (${\text{\xmark}}$)  & \textit{speed}, \textit{design}, \textit{simplicity} & [\textit{speed}]$_{\texttt{POS}}$, [\textit{design}]$_{\texttt{POS}}$, [\textit{simplicity}]$_{\texttt{POS}}$\\ \hline
        6. The \textbf{[}\textcolor{red}{\textit{battery life}}\textbf{]}$_{\texttt{POS}}$ is excellent , the \textbf{[}\textcolor{red}{\textit{display}}\textbf{]}$_{\texttt{POS}}$ is excellent and \textbf{[}\textcolor{red}{\textit{downloading apps}}\textbf{]}$_{\texttt{POS}}$ is a breeze . & \textit{battery} (${\text{\xmark}}$), \textit{display}, \textit{apps} (${\text{\xmark}}$)  & \textbf{[}\textit{battery}\textbf{]}$_{\texttt{POS}}$ (${\text{\xmark}}$), \textbf{[}\textit{display}\textbf{]}$_{\texttt{POS}}$, \textbf{[}\textit{apps}\textbf{]}$_{\texttt{POS}}$ (${\text{\xmark}}$)  &   \textit{battery} (${\text{\xmark}}$), \textit{display}, \textit{apps} (${\text{\xmark}}$)  & \textbf{[}\textit{battery}\textbf{]}$_{\texttt{POS}}$ (${\text{\xmark}}$), \textbf{[}\textit{display}\textbf{]}$_{\texttt{POS}}$, \textbf{[}\textit{apps}\textbf{]}$_{\texttt{POS}}$ (${\text{\xmark}}$)   &  \textit{battery life}, \textit{display}, \textit{downloading apps} & \textbf{[}\textit{battery life}\textbf{]}$_{\texttt{POS}}$, \textbf{[}\textit{display}\textbf{]}$_{\texttt{POS}}$, \textbf{[}\textit{downloading apps}\textbf{]}$_{\texttt{POS}}$\\ \hline
        \Xhline{3\arrayrulewidth}
    \end{tabular}}
    \caption{Case analysis for the $\mathbb{R}$$\rightarrow$$\mathbb{L}$ pair. Note that we only show the sentiment part of the unified labels (i.e., \texttt{POS}, \texttt{NEG}, and \texttt{NEU}) and use brackets to indicate the boundary. The marker ${\text{\xmark}}$ denotes an incorrect prediction. }
    \label{tab:case_study}
    \vspace{-2mm}
\end{table*}

Note that, the AD-AL, AD-SAL (\textbf{Full model}) and ADS-SAL all use the same architecture as the Base Model+DMI.
Based on the Table~\ref{tab:ablation_results}, we have the following observations to give us evidences:

\begin{itemize}
\item \textbf{No DMI} v.s. \textbf{DMI}: Base Model+DMI outperforms the Base Model (SO) by 9.80\% and 9.70\% Micro-F1 on average for the AD and ADS tasks, respectively. This demonstrates that the original word hidden representations (LSTM$^\mathcal{B}$ hidden states) are not suitable for transfer. Thus, we need to resort to the correlation vectors inferred by the DMI that models the transferable latent relations between aspect and opinion words.
\item \textbf{No SAL} v.s. \textbf{SAL}: AD-SAL significantly and consistently exceeds Base Model+DMI by 6.46\% and 4.92\% Micro-F1 on average for the AD and ADS tasks, respectively. Without any adaptation, the captured relations by the DMI may not work well across domains, while the proposed SAL method can effectively align these latent relations to be domain-invariant.
\item  \textbf{No Selectivity} v.s. \textbf{Selectivity}: AD-SAL outperforms AD-AL by 5.18\% and 4.11\% Micro-F1 on average for the AD and ADS tasks, respectively. This proves the necessity to conduct the selective alignment. The SAL method can dynamically learn to control an alignment weight for each word to achieve a local semantic alignment, which captures a better domain-invariant feature space than pure adversarial learning that treats all words equally for the fine-grained adaptation.
\item \textbf{Low-level} v.s. \textbf{High-level}: AD-SAL exceeds ADS-SAL by 2.31\% and 1.94\% Micro-F1 on average for the AD and ADS tasks, respectively. The label space of the {\it unified tags} for the high-level ADS task is more complicated than that of the {\it aspect boundary tag} for the low-level AD task. This gives the evidence that low-level neural features are more easily to transfer than high-level features.
\end{itemize}

\begin{figure}[t]
\centering
\includegraphics[width=1.0\columnwidth]{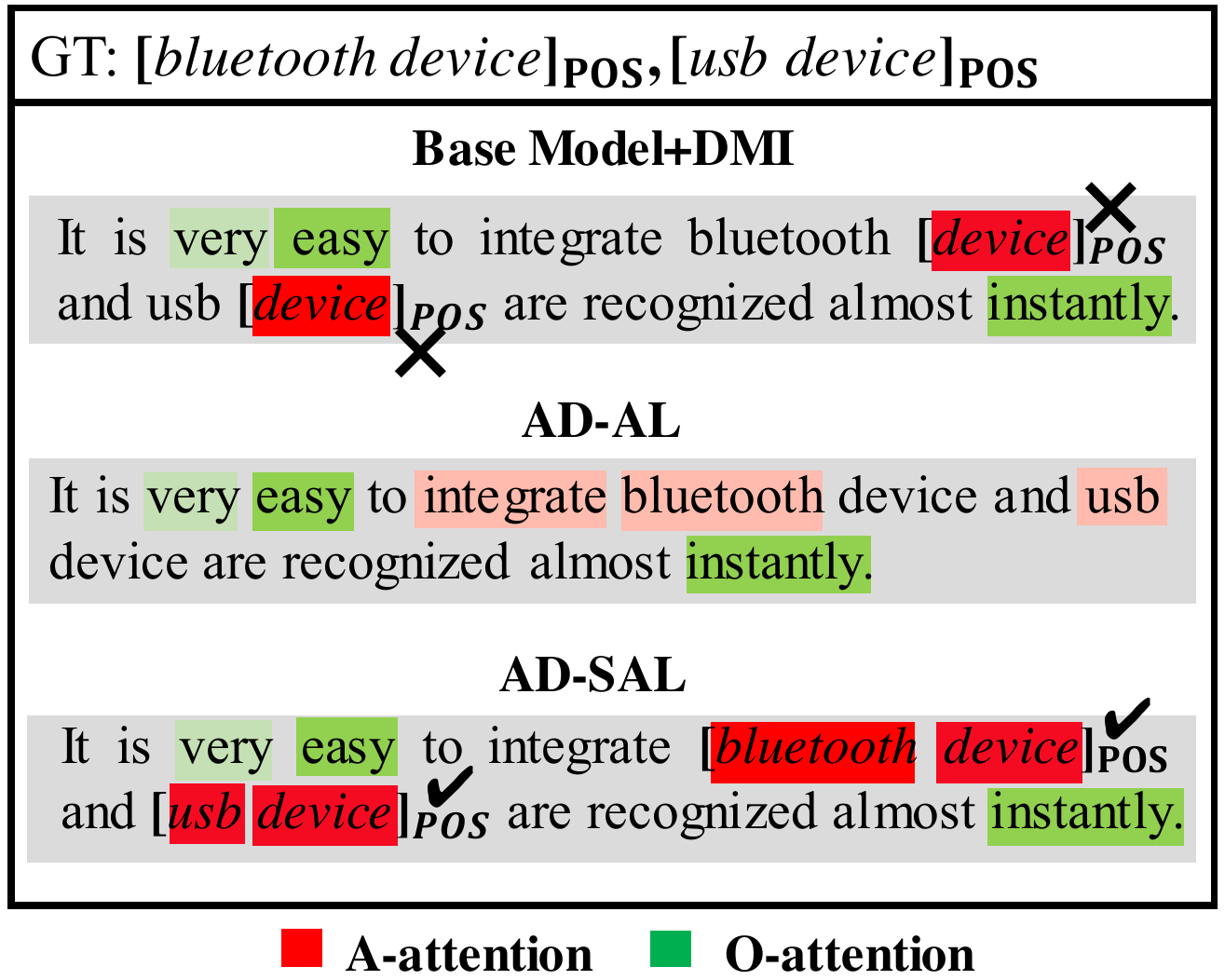}
\caption{Visualization of attention for the $\mathbb{R}$$\rightarrow$$\mathbb{L}$ pair.}\label{fig:attention}
\vspace{-2mm}
\end{figure}

\subsection{Case Analysis}
As illustrated in Table~\ref{tab:case_study}, we perform case analysis of the $\mathbb{R}$$\rightarrow$$\mathbb{L}$ pair for the Base model+DMI, AD-AL, and the full model AD-SAL to demonstrate the necessity to conduct selective alignment for the fine-grained adaptation. The Base model+DMI can identify some domain-specific aspect words (e.g., {\it battery}, {\it ports}) without any supervision from the target domain. This is because the DMI can infer relational representations that capture some common latent relations between aspect and opinion words. However, it still cannot completely capture the multi-word aspect terms (e.g. {\it bluetooth device}, {\it battery life}) and sometimes it totally ignores them (e.g. {\it window 7}). The AD-AL performs pure adversarial learning for aligning all words in a sentence. Even though the domain adaptation method is adopted, it does not yield significant improvements and sometimes it becomes worse. For example, the AD-AL cannot even identify aspect words that can be captured by the Base Model+DMI (e.g., {\it ports}, {\it devices}), or wrongly identifies some non-aspect words (e.g., {\it pc}). The reason is that pure adversarial learning treats all the words equally for the alignment, which may bring in noises of uninformative words into the shared feature space. To solve that, the full model AD-SAL performs a local semantic alignment to dynamically focus on aligning aspect words that contribute more to the domain-invariant feature space. Hence, AD-SAL model can precisely and completely identity all the aspect terms and make correct unified tag predictions. 

Moreover, in Figure~\ref{fig:attention}, we visualize the attentions from these models, where deeper colors denote larger weights. Compared with the Base model+DMI, the full model AD-SAL can precisely attend the complete aspect words from the target domain (A-attention), i.e., {\it bluetooth device} and {\it usb device}, and make correct predictions, while the AD-AL cannot achieve that. The AD-AL can only align all the words equally, which hinders the model to attend the aspect words, while the A-attention of the AD-SAL model can be used for both discriminative word tags predictions and acting as a learnable alignment weight for each word. This shows that the proposed SAL method can learn to align important aspect words to improve the transferability of the model for the fine-grained adaptation.

\section{Related Works}
E2E-ABSA can be broken into two sub-tasks, namely, aspect detection and aspect sentiment classification. The aspect detection aims to extract the aspect terms mentioned in the text and it has been actively studied~\cite{qiu2011opinion,liu2015fine,poria2016aspect,wang2016recursive,he2017unsupervised,wang2017coupled,majumder2018iarm,li2018aspect,xu2018double}. The aspect sentiment classification is to predict the sentiment polarities of the given aspect terms and has also received a lot of attention recently~\cite{dong2014adaptive,tang2016aspect,wang2016attention,ma2017interactive,chen2017recurrent,ma2018targeted,he2018exploiting,li2018transformation,li2019exploiting}. 
For practical applications, a typical way is to pipeline these two sub-tasks together, which becomes ineffective due to the accumulated errors across tasks. These two sub-tasks have strong couplings and thus a unified formulation~\cite{mitchell2013open,zhang2015neural,li2019unified} to handle them together in an end-to-end manner becomes a more promising direction. Despite its importance, existing studies are only exploring the performance in a single domain, while ignoring the transferability across domains. To address this problem, unsupervised domain adaptation methods can be applied. While existing methods focus on traditional cross-domain sentiment classification to learn shared representations for sentences or documents, including pivot-based methods~\cite{blitzer2007biographies,pan2010cross,bollegala2013cross,yu2016learning}, auto-encoders~\cite{glorot2011domain,chen2012marginalized,zhou2016bi}, domain adversarial networks~\cite{ganin2016domain,li2017end,li2018hatn}, or semi-supervised methods~\cite{he2018adaptive}. Due to the difficulties in fine-grained adaptation, there exist very few methods for cross-domain aspect extraction~\cite{jakob2010extracting,ding2017recurrent}, which acts as a sub-task of E2E-ABSA, or aspect and opinion co-extraction~\cite{li2012cross,wang2018recursive} that focuses on detecting aspect and opinion words, while E2E-ABSA needs to analyze more complicated correspondences between them. Besides, those methods can only rely on general syntactic relations between aspect and opinion words to bridge the domains.
Different from them, our method leverages the attention mechanism~\cite{bahdanau2014neural,sukhbaatar2015end,shen2017disan,shen2019tensorized} as a dynamic selector to automatically achieve the selective alignment.

\section{Conclusion}
The effectiveness of supervised methods for E2E-ABSA is limited due to the data scarcity. Our wok is the first attempt to resolve cross-domain E2E-ABSA by leveraging knowledge from other related domains to enhance the sequence learning of the target domain. Extensive experiments show the effectiveness of the proposed SAL method. Ablation studies also prove the necessity to perform selective alignment. In the future, the proposed SAL method can be potentially extended to other domain adaptation methods and applied to more general sequence labeling tasks including named entity recognition~\cite{zhou2019dual_c}, part-of-speech tagging~\cite{zhou2019dual_j}, etc.

\section*{Acknowledgement}
We thank the support of Hong Kong CERG grants (16209715 \& 16244616) and NSFC 61673202.

\bibliography{san}
\bibliographystyle{acl_natbib}

\end{document}